\documentclass[a4paper,fleqn]{cas-dc}

\usepackage{blindtext}
\usepackage{bm}
\usepackage[numbers]{natbib}
\setcitestyle{sort&compress}
\usepackage{graphicx}
\usepackage{float}
\usepackage{ragged2e}

\usepackage{hyperref}
\hypersetup{
    colorlinks=true,
    linkcolor=blue,
    filecolor=blue,      
    urlcolor=blue,
    citecolor=blue,
}

\usepackage{geometry}
\usepackage{microtype}
\usepackage{amssymb}
\usepackage{amsthm}
\usepackage{amsmath}
\usepackage{caption}
\makeatletter 
\def\@makecaption#1#2{%
  \vskip\abovecaptionskip
  \sbox\@tempboxa{#1 #2}%
    {\bfseries #1} #2\par
  \vskip\belowcaptionskip}
\makeatother
\usepackage{booktabs}
\usepackage[justification=centering]{caption}

\begin{document}
\begin{sloppypar}

\let\printorcid\relax
\def\floatpagepagefraction{1}
\def\textpagefraction{.001}
\shorttitle{}
 \shortauthors{H. Sal.}

\title [mode = title]{Disentangling Heterogeneous Knowledge Concept Embedding for Cognitive Diagnosis on Untested Knowledge}                      

\author[]{Miao Zhang}

\author[]{Ziming Wang}

\author[]{Runtian Xing}

\author[]{Kui Xiao}
\cormark[1]

\author[]{Zhifei Li}

\author[]{Yan Zhang}

\author[]{Chang Tang}

\cortext[cor1]{\ Corresponding author.\\\indent \ \ \ \ \ \  E-mail addresses: xiaokui@hubu.edu.cn (K. Xiao)}

\begin{abstract}
Cognitive diagnosis is a fundamental and critical task in learning assessment, which aims to infer students' proficiency on knowledge concepts from their response logs. Current works assume each knowledge concept will certainly be tested and covered by multiple exercises. However, whether online or offline courses, it's hardly feasible to completely cover all knowledge concepts in several exercises. Restricted tests lead to undiscovered knowledge deficits, especially untested knowledge concepts(UKCs). In this paper, we propose a novel framework for Cognitive Diagnosis called \underline{Dis}entangling Heterogeneous \underline{K}nowledge \underline{C}ognitive \underline{D}iagnosis(DisKCD) on untested knowledge. Specifically, we leverage course grades, exercise questions, and learning resources to learn the potential representations of students, exercises, and knowledge concepts. In particular, knowledge concepts are disentangled into tested and untested based on the limiting actual exercises. We construct a heterogeneous relation graph network via students, exercises, tested knowledge concepts(TKCs), and UKCs. Then, through a hierarchical heterogeneous message-passing mechanism, the fine-grained relations are incorporated into the embeddings of the entities. Finally, the embeddings will be applied to multiple existing cognitive diagnosis models to infer students' proficiency on UKCs. Experimental results on real-world datasets show that the proposed model can effectively improve the performance of the task of diagnosing students' proficiency on UKCs. Our code is available at \url{https://github.com/Hubuers/DisKCD}.
\end{abstract}


\maketitle

\section{Introduction}
\label{sec_intro}
Cognitive diagnosis is an important component of intelligent tutoring systems \citep{DBLP:conf/ijcai/Liu21, DBLP:conf/www/AndersonHKL14, DBLP:journals/fcsc/LiuZWYL23}, which aim to assess students’ proficiency in specific knowledge concepts. Cognitive diagnosis models (CDMs) analyze students' interaction history, including correct and incorrect responses in exercise logs, to infer their cognitive states. Presently, cognitive diagnosis outcomes are integrated into various facets of intelligent tutoring, such as exercise recommendation \citep{DBLP:conf/kdd/0038HLBW23, DBLP:journals/peerj-cs/MaoL23} and adaptive testing \citep{DBLP:conf/nips/ZhuangLZHHPCWL23}. As the array of learning materials expands and students' learning preferences vary more widely, cognitive diagnosis models are becoming increasingly important. They furnish instructors with potent psychometric instruments for educational appraisal, enhancing teaching and learning support.

In cognitive diagnosis, researchers propose various CDMs, such as Item Response Theory (IRT) \citep{lord1952theory}, Multidimensional IRT (MIRT) \citep{Reckase2009}, and the Deterministic Inputs, Noisy AND gate model (DINA) \citep{de2009dina}. These models employ artificially designed linear functions to simulate student-practice interactions. However, in recent years, due to the proliferation of neural networks and deep learning technologies, researchers have introduced novel cognitive diagnosis models leveraging neural networks, such as Neural Cognitive Diagnosis (NeuralCD) \citep{DBLP:conf/aaai/WangLCHCYHW20}, and models based on graph neural networks like Relation map driven Cognitive Diagnosis (RCD) \citep{DBLP:conf/sigir/Gao0HYBWM0021}. These advanced models have achieved remarkable results in the field of cognitive diagnosis.

\begin{figure*}
  \centerline{\includegraphics[width=1\textwidth]{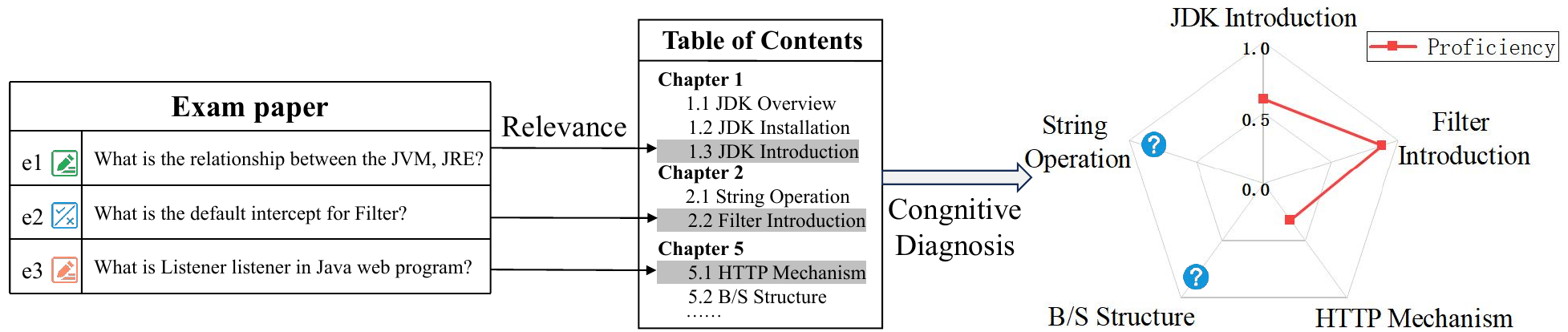}}
  \caption{Examples of cognitive diagnosis. Students' mastery of tested knowledge components is often emphasized, while their grasp of untested knowledge components is usually overlooked.}\label{FIG:1}
\end{figure*}

Despite these advancements, educational assessments still face practical challenges. Researchers typically assume that each knowledge concept is tested through multiple exercises, focusing on students’ mastery of tested knowledge concepts (TKCs). However, as shown in Fig~\ref{FIG:1}, the final exam paper for the course contains a limited number of exercises, covering only a fraction of the knowledge concepts of the course, such as chapters 1.3, 2.2, and 5.1. Instructors can assess student understanding of TKCs in the final exam \citep{DBLP:conf/nips/ChenWLCZHW23}. However, constraints of time and exam scope make it difficult to cover all knowledge concepts through a limited number of exercises, leading to many concepts being insufficiently tested. This poses a significant challenge in diagnosing untested knowledge concepts (UKCs), as these untested concepts often represent potential "blind spots" in students’ learning. Although these concepts are not directly assessed, they can still profoundly affect students’ overall performance and knowledge structure. Relying solely on TKCs evaluations may hinder teachers from gaining a comprehensive understanding of students’ learning progress and result in missed opportunities for timely intervention and personalized guidance. Therefore, effectively diagnosing students’ mastery of UKCs is crucial. It not only helps identify knowledge gaps but also provides essential data for designing personalized learning paths. Through such comprehensive diagnosis, educators can gain deeper learning insights, address untested knowledge gaps more precisely, and ultimately enhance students’ overall development and learning outcomes.

In this paper, we propose a novel cognitive diagnosis framework aimed at effectively diagnosing students' mastery of untested knowledge concepts (UKCs), which traditional assessments often overlook. Recognizing the limitations of existing methods, our approach first disentangles the knowledge concepts into two categories: tested knowledge concepts (TKCs), and untested knowledge concepts(UKCs). We then construct a heterogeneous relation graph network, comprising entities such as students, exercises, TKCs, and UKCs, to capture the intricate relationships among them. This network facilitates the integration of fine-grained information through a hierarchical heterogeneous message-passing mechanism, which enhances the entity embeddings with nuanced relational data. Finally, by applying these enhanced embeddings to the interaction functions of various diagnostic models, we can predict students' performance on untested exercises, thereby assessing their mastery of UKCs and supporting personalized learning and targeted interventions.

Our key contributions are summarized as follows:
\begin{itemize}
    \item We propose a novel cognitive diagnosis framework to assess students' proficiency on UKCs, which effectively addresses the challenge of difficulty in comprehensively examining traditional cognitive diagnosis.
    
    \item We present a heterogeneous relation aware network for students, exercises, TKCs, and UKCs to ensure that the embeddings are highly sensitive to students’ proficiency and effectively disentangling between UKCs and TKCs.
    
    \item We create an authentic dataset from two university courses, facilitating cognitive diagnosis research on UKCs. Diverging from previous datasets on Intelligent Tutoring Systems platforms, our dataset is meticulously curated from genuine data extracted from final exams of university courses, which are better suited to address the complexities of evaluating students’ proficiency in paper-and-pencil test contexts.
\end{itemize}

\section{Related Work}
\label{sec_rel}

\subsection{Cognitive Modeling}
\label{ssec_CognitiveDiagnosis}

Cognitive modeling is a comprehensive approach to educational assessment, involving two main tasks: Cognitive Diagnosis \citep{DBLP:conf/aaai/WangZYX024,DBLP:conf/aaai/MaWZYZ024,DBLP:journals/isci/ZhaoMWHC24} and Knowledge Tracing \citep{DBLP:conf/kdd/0001HL20, DBLP:conf/mm/SunYLLLS23, DBLP:conf/aaai/Wang0Z23}. These two tasks serve complementary purposes. Knowledge tracing focuses on monitoring changes in student performance to predict their responses to future questions, thus it excels at capturing the ongoing dynamics of students' state of knowledge as they answer questions. For instance, recent works have explored deep learning approaches in knowledge tracing, such as using recurrent neural networks (RNN) and attention-based mechanisms to model students’ evolving knowledge states more effectively. However, knowledge tracing does not directly evaluate how well students grasp concepts throughout their learning process. 

On the other hand, in cognitive diagnosis, early models like Item Response Theory (IRT) \citep{lord1952theory} laid the groundwork by using manually crafted functions to predict students' answer accuracy, emphasizing simplicity and interpretability. Building on this foundation, Multidimensional IRT (MIRT) \citep{Reckase2009} extended IRT’s capabilities by incorporating multiple cognitive dimensions, providing a more detailed view of student knowledge states . Despite these enhancements, traditional models still rely heavily on well-designed functions, limiting their adaptability in complex educational settings \citep{DBLP:conf/edm/TsutsumiKU21, DBLP:journals/fgcs/GaoZLZZ22}. To address this issue, Neural Cognitive Diagnosis (NeuralCD) \citep{DBLP:conf/aaai/WangLCHCYHW20, DBLP:journals/access/JiangW20d, DBLP:journals/tkde/AbdelrahmanW23} introduced deep learning to cognitive diagnosis, leveraging multilayer perceptrons (MLPs) to capture non-linear relationships between students' knowledge states and performance. Expanding this further, Relation map driven Cognitive Diagnosis (RCD) \citep{DBLP:conf/sigir/Gao0HYBWM0021} utilized graph neural networks to analyze the interactions among students, exercises, and knowledge concepts, enhancing diagnostic accuracy for complex, large-scale data. Meanwhile, models like the Deterministic Inputs, Noisy AND gate (DINA) \citep{de2009dina} used discrete values to represent mastery and addressed issues such as slipping and guessing errors. Building upon DINA, the Fuzzy Cognitive Diagnosis Model (FuzzyCDM) \citep{DBLP:journals/tist/LiuWCXSCH18} introduced fuzzy logic, allowing for a more nuanced evaluation of student proficiency when knowledge mastery isn’t absolute.

In summary, different cognitive modeling approaches possess distinct strengths. Knowledge tracing is effective at capturing the dynamic shifts in students' knowledge over time, whereas cognitive diagnosis offers a detailed, static evaluation of their current understanding.

\subsection{Knowledge Coverage Problem}
\label{ssec_KnowledgeCoverage}
Insufficient knowledge coverage in exercises poses a major challenge for cognitive assessment tasks\citep{DBLP:journals/tkde/LiuHYCXSH21, DBLP:journals/corr/abs-2312-10110}. As educational settings increasingly rely on standardized tests and digital platforms, exercises often fail to cover the full range of knowledge concepts. Current CDMs focus on a limited set of exercises, which can result in critical knowledge gaps by emphasizing frequently tested concepts while overlooking less common or complex areas. Sparse student responses add to this difficulty, especially in large-scale automated assessment systems. To overcome these challenges, adaptive testing systems that can assess students’ understanding across both covered and uncovered knowledge concepts are essential, particularly as educational content continues to diversify.

Recent advancements in cognitive diagnosis models aim to address these gaps through innovative approaches. For instance, the Inductive Cognitive Diagnosis Model (ICDM) \citep{DBLP:conf/www/LiuSQZ24} for web-based online intelligent education systems leverages a student-centered graph (SCG) to provide fast, inductive diagnosis. By aggregating mastery levels from similar students, ICDM avoids retraining and is particularly efficient for new students, thereby enhancing scalability in open learning environments. The Structure-Aware Inductive Knowledge Tracing Model (SINKT) \citep{DBLP:journals/corr/abs-2407-01245} further explores this by addressing data sparsity and cold-start issues. SINKT employs both semantic and structural encoders to capture nuanced relationships within heterogeneous graphs of concepts and exercises, thereby improving the accuracy of knowledge tracing. In addition, the Knowledge-Sensed Cognitive Diagnosis (KSCD) framework \citep{DBLP:conf/cikm/MaLWZC0Z22} has been introduced to tackle the complexities of cognitive diagnosis on intelligent education platforms. KSCD aims to infer students' mastery over both observed and non-interactive knowledge concepts by learning intrinsic relationships among knowledge concepts directly from student response logs. It achieves this by projecting students, exercises, and knowledge concepts into embedding matrices, capturing these intrinsic relationships within the knowledge embedding representation. This embedding-based approach allows KSCD to infer mastery over knowledge concepts not directly covered in exercises, offering a more comprehensive diagnostic assessment.

In addition, zero-shot cognitive diagnosis at the domain level is also a recent hot topic, such as newly TechCD \citep{DBLP:conf/sigir/GaoWLWLYZL023} and Zero-1-to-3 \citep{DBLP:conf/aaai/GaoLWYBGYZLH24}. Zero-shot learning approaches offer exciting possibilities for diagnosing student proficiency in new or untested domains, by leveraging knowledge transfer from related tasks or domains. This could potentially reduce the need for extensive domain-specific training data, which is often a major bottleneck in the development of robust cognitive diagnosis models. Unlike these works that explore cognitive diagnosis across different domains, our approach specifically addresses the challenge of assessing UKCs within examinations. While KSCD also aims to infer students' mastery over non-interactive knowledge concepts, it does so by embedding knowledge concepts directly from student response logs, focusing on relational embeddings within a single vector space. In contrast, DisKCD uniquely employs a heterogeneous relational graph network to explicitly differentiate between TKCs and UKCs. This graph-based structure incorporates course grades, exercises, and learning resources, allowing for fine-grained message passing among students, exercises, TKCs, and UKCs. Through this hierarchical heterogeneous message-passing mechanism, DisKCD captures complex relationships and potential dependencies across multiple entities, which is essential for accurately diagnosing proficiency on UKCs that are not directly covered in exercises. To the best of our knowledge, this is the first time that a relational graph network has been used to capture relationships between knowledge concepts to assess students' mastery of UKCs. This pioneering approach opens new avenues for research in adaptive learning systems and personalized education, where understanding the full spectrum of student knowledge is crucial for effective intervention and support.

\section{Preliminaries}

\label{sec_Preliminaries}

\subsection{Cognitive Diagnosis Model}
\label{ssec_CognitiveDiagnosisModel}

We first outline the general form of CDMs, which typically consist of three fundamental components: students, exercises, and knowledge concepts. Most CDMs predict how well students will perform in practicing exercises, indirectly quantifying students' mastery of knowledge. The diagnosis core essentially involves modeling student-exercise-concept interactions by predicting students' performance during practice sessions, as illustrated below:
\begin{equation}\label{eq:fcdm}
    {\hat{y}}_{uv}=F_{CDM}(u,v,K),
\end{equation}

Here, $u$ denotes the traits of a student, $v$ represents the traits of exercise, and $K$ is the knowledge concept space of a domain. The primary aim of CDMs is to minimize the disparity between the anticipated outcome ${\hat{y}}_{uv}$ and the actual score $y_{uv}$ of the student's response. $F_{CDM}(\cdot)$ is a diagnostic tool for predicting the response outcome. 

Interpretability is significantly important for cognitive diagnosis for improving teaching and learning \citep{DBLP:conf/xai/ChaushiSCA23}. CDMs typically follow the monotonicity assumption in the literature \citep{Reckase2009}: "The probability of correct response to the exercise is monotonically increasing at any dimension of the student’s knowledge proficiency." In pursuit of the assumption, CDMs enforce $\frac{\partial F}{\partial u}>0$ throughout the training phase.

\begin{figure*}
  \centerline{\includegraphics[width=1\textwidth]{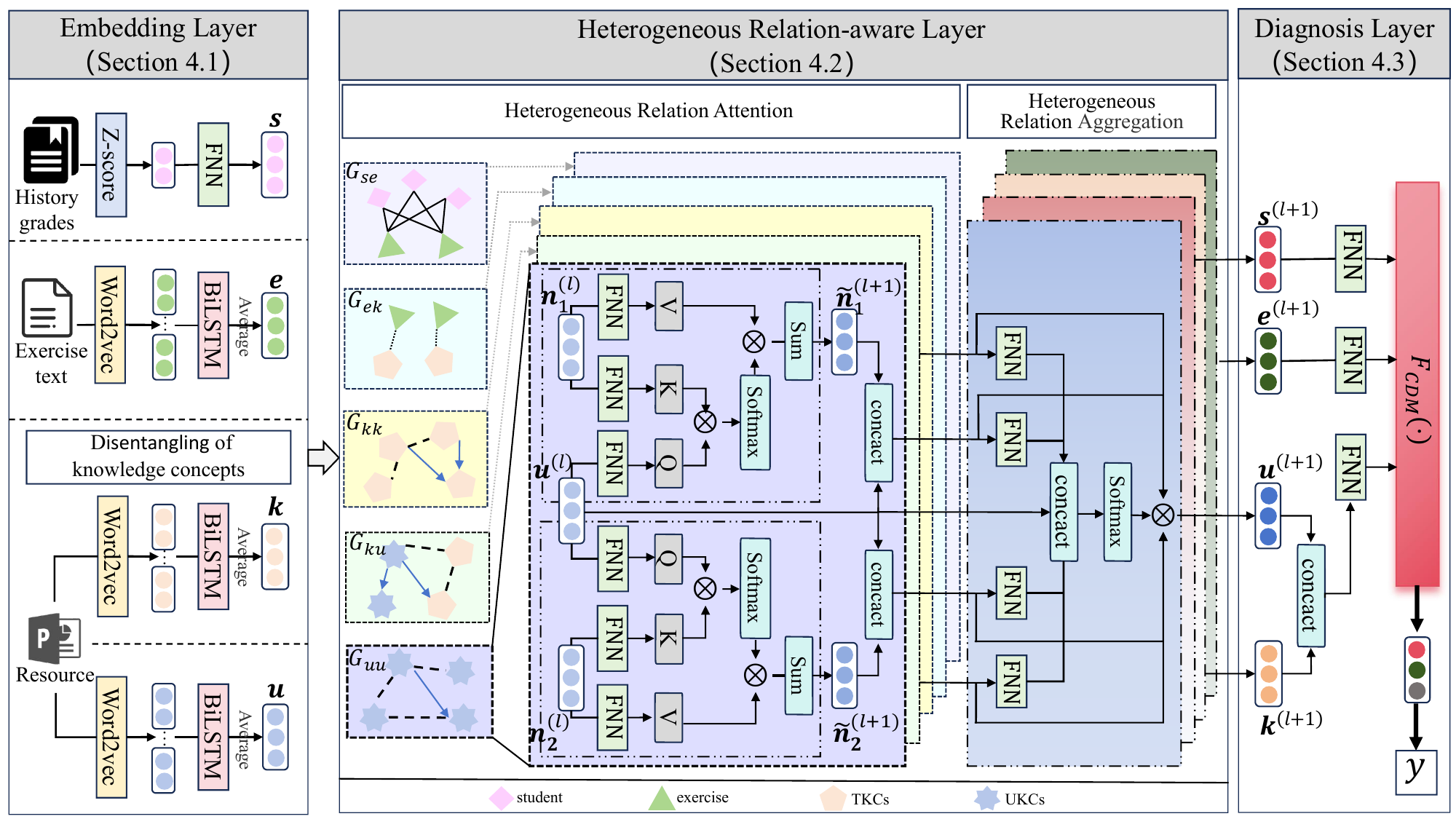}}
  \caption{Overview of DisKCD. (1) The left side shows the disentangling of knowledge concepts and the original embedding of students and exercises. (2) The middle part shows the details of the heterogeneous relation-aware layer where fine-grained relations are merged into entity embeddings. (3) The right part shows the diagnostic process of DisKCD framework.}\label{FIG:2}
\end{figure*}

\subsection{Problem Setup}
\label{ssec_ProblemSetup}

In the cognitive diagnosis task involving UKCs, we define several terms. Let $S=\left\{s_1,\ s_2,\ ...,\ s_N\right\}$ represent the set of students, and $E=\left\{e_1,\ e_2,\ ...,\ e_M\right\}$ denote the exercises within a course exam. $K_{tested}$ signifies the knowledge concepts corresponding to these exercises, and $K_{untested}$ denote the knowledge concepts not covered in the exam. The union $K=K_{tested}\cup K_{untested}$ encompasses all knowledge concepts in the course, with the constraint $K_{tested}\cap K_{untested}=\Phi$. Let $R_{ht}=\left\{(k_h,k_t,r_{ht})|k_h,k_t\in K,\ r_{ht}\in R_k\right\}$ denotes the set of triples of knowledge concepts, where $R_k$ contains all forms of relations among knowledge concepts. The present study, inspired by RCD \citep{DBLP:conf/sigir/Gao0HYBWM0021}, specifically employs two types of relations: similarity and prerequisite relations.

On the other hand, the observed interaction logs are a triple set $R_{se}=\left\{(s_u,e_v,y_{uv})|s_u\in S,e_v\in E,y_{uv}\in{0,1}\right\}$, where $y_{uv}$ indicates the response of student $s_u$ to exercise $e_v$. A correct response is denoted by $y_{uv}=1$, while an incorrect response is indicated by $y_{uv}=0$. Additionally, we have an expert-annotated Q-matrix \citep{tatsuoka2012architecture}, where $Q_{ij}=1$ signifies that exercise $e_i$ is associated with knowledge concept $k_j$, otherwise $Q_{ij}=0$.

It is important to note that the resources available for diagnosing students differ between courses on Intelligent Tutoring System (ITS) platforms and offline courses. Generally, offline courses have access to a broader range of information related to students, exercises, and concepts. However, in online courses, ITS platforms typically only provide IDs for students, exercises, and concepts. Therefore, this study makes different assumptions based on the type of course: for online course datasets, we assume that only the IDs of students, exercises, and concepts are known; whereas for offline course datasets, we assume that additional information, such as students' historical performance, exercise texts, and concept definitions, is available. This difference influences the methods used for generating initial embeddings for students, exercises, and concepts.

\textbf{PROBLEM DEFINITION.} Given students' exercising response logs $R_{se}$ on TKCs, the set of triples of knowledge concepts $R_{ht}$, as well as the $Q-matrix$, our goal is to indirectly assess students' proficiency in UKCs by predicting their performance on exercises related to these UKCs.

\subsection{Heterogeneous Relation Graphs}
\label{ssec_HeterogeneousRelationGraph}

Depending on dynamic interactions between students, exercises, and knowledge concepts, we define diverse heterogeneous relation subgraph.

\textbf{Student-Exercise interaction subgraph.}
The interaction relations between students and exercises listed on test paper collectively form the student-exercise interaction subgraph. This subgraph is bidirectional and can be represented as $G_{se}(S\cup E, r_{se})$, where $r_{se}$ denotes the set of interactions between students and the exercises. If $r_{s_u\leftrightarrow e_v}=1$, it indicates that student $s_u$ attempted exercise $e_v$. 

\textbf{Exercise-TKC correlation subgraph.}
The incorporation connection between TKCs and exercises collectively forms the Exercise-TKC correlation subgraph, denoted as $G_{ek}(E\cup K_{tested}, r_{ek})$. Here, all $r_{ek}$ derive from the Q-matrix. If $r_{e_v\leftrightarrow k_w}$ within this set equals 1, it indicates that exercise $e_v$ encompasses the knowledge concept $k_w$.

\textbf{TKC-TKC dependency subgraph.}
The connections among the TKCs collectively form the TKC-TKC dependency subgraph, denoted as $G_{kk}(K_{tested}, r_{kk})$. We solely contemplate the similarity and prerequisite relations. The similarity relation allows the propagation of students' cognitive states between concepts in both directions. Conversely, prerequisite relations function unidirectionally, specifically transferring from prerequisite concepts to successor concepts.

\textbf{TKC-UKC dependency subgraph.}
The connections among TKCs and UKCs, denoted as $G_{ku}(K_{tested}\cup K_{untested}, r_{ku})$. Just like $r_{kk}$, $r_{ku}$ encompasses the array of connections between TKCs and UKCs, encompassing similarities and prerequisites. Moreover, the nodes linked by a relation must embody a TKC and a UKC.

\textbf{UKC-UKC dependency subgraph.}
The connections among all UKCs collectively form the UKC-UKC dependency subgraph, represented as $G_{uu}(K_{untested}, r_{uu})$. Much like $r_{kk}$,$r_{uu}$ denotes the array of relations among UKCs. Likewise, the connections between concept nodes here solely comprise similarity and prerequisite relations.

\section{Methodology}
\label{sec_Methodology}

This paper presents a DisKCD framework for UKCs, which aims to address the challenge of assessing students' mastery of UKCs when the concepts already examined in their assignments and tests are limited. First, in the embedding layer, we disentangle knowledge concepts to TKCs and UKCs. Then, in the heterogeneous relation aware layer, the fine-grained relations are incorporated into the entity embeddings. Finally, embeddings are applied to CDMs to infer students' proficiency on UKCs.The overall framework of DisKCD is shown in Fig~\ref{FIG:2}.

\subsection{Embedding Layer}
\label{sec_EmbeddingLayer}

The embedding layer encodes students, test questions, and knowledge concepts into initial embeddings that capture essential semantic details. Student embeddings are based on historical performance data, which reflects their prior knowledge states. Test questions are embedded with features that convey both content and difficulty, while knowledge concepts are disentangled into tested and untested initial embeddings, allowing the model to distinguish between these areas effectively. This setup provides a robust foundation for subsequent cognitive diagnosis tasks, generating context-rich vectors that prepare the model for heterogeneous relation-aware processing, enabling a more nuanced cognitive diagnosis.

\subsubsection{Student Embedding}
As previously mentioned, students' historical performance is known in offline courses, which helps to comprehensively assess their initial abilities. Specifically, we gather and normalize their grades using the standard score (z-score), which is the number of standard deviations above or below the mean value of what is being observed or measured \citep{curtis2016mystery}. The conversion operation is
\begin{equation}
    z_c^{(u)}=\frac{X_c^{(u)}-\mu_c}{\sigma_c},
\end{equation}
where $X_c^{(u)}$ is the grade of student $s_u$ in course $c$, $\mu_c$ denotes the average score for $c$. $\sigma_c$ stands for the standard deviation of all scores within the class for this particular course.
For each student, we obtain an original embedding via a fully connected neural network:
\begin{equation}
    \boldsymbol{s}=FNN(z_1,\ z_2,\ ...,\ z_C).
\end{equation}
Here, $\boldsymbol{s}\in\mathbb{R}^{d_k\times1}$ and $d_k$ denotes the number of knowledge concepts in the current course. $FNN(\cdot)$ represents a fully connected network. $C$ represents the number of historical courses. The embedding $s$ incorporates all of the student’s historical performance information, reflecting their initial ability before starting the current course.

For online courses, since the student's historical performance information is unknown, we generate the student's initial embedding by multiplying the student's one-hot vector with a learnable matrix.

\subsubsection{Exercise embedding}
For offline courses, the written content of the exercises is known, and we construct the exercise embeddings based on this content. Specifically, we feed the word2vec \citep{DBLP:conf/nips/MikolovSCCD13} sequence $(\boldsymbol{w}_1,\ \boldsymbol{w}_2,\ ...,\ \boldsymbol{w}_T)$ of exercise into each time step of Bi-LSTM \citep{DBLP:journals/nn/GravesS05}, where $T$ represents the total number of words in the exercise. Subsequently, we concatenate the hidden state by the forward direction $(\overleftarrow{\boldsymbol{h}}_1,\ \overleftarrow{\boldsymbol{h}}_2,\ ...,\ \overleftarrow{\boldsymbol{h}}_T)$ and the backward direction $(\overrightarrow{\boldsymbol{h}}_1,\ \overrightarrow{\boldsymbol{h}}_2,\ ...,\ \overrightarrow{\boldsymbol{h}}_T)$, aligning them by position, and then yielding
\begin{equation} \label{eq:lstm}
    \boldsymbol{h}_i=[\overleftarrow{\boldsymbol{h}}_i,\overrightarrow{\boldsymbol{h}}_i], 
\end{equation}
where $[,]$ denotes the concatenation operation. 
Finally, the exercise vector $\boldsymbol{e}\in\mathbb{R}^{d_k\times1}$ is updated by averaging operation and a fully connected layer:
\begin{equation}\label{eq:fnn}
    \boldsymbol{e}=FNN(AVG(\boldsymbol{h}_1,\ \boldsymbol{h}_2,\ ...,\ \boldsymbol{h}_T)). 
\end{equation}

Similarly, for online courses, if the exercise content is not provided, we generate the initial exercise embeddings by multiplying the one-hot vector with a learnable matrix.

\subsubsection{Concept embedding}
For offline courses, the definitions of concepts are known, derived from concept-related learning resources (such as course slides). We use the textual information of these concepts to generate their initial embeddings, enhancing the semantic representation of the knowledge concepts. The embedding construction process for the knowledge concept is the same as Eq.\ref{eq:lstm} and Eq.\ref{eq:fnn}. Specifically, we denote the embedding of TKC and UKC as $\boldsymbol{k}\in\mathbb{R}^{d_k\times1}$ and $\boldsymbol{u}\in\mathbb{R}^{d_k\times1}$ respectively.

Similarly, for online courses, if the content of the concept definitions is unavailable, we generate the initial concept embeddings by multiplying the one-hot vector with a learnable matrix.

\subsection{Heterogeneous Relation-aware Layer}
\label{ssec_Heterogeneous_relation_aware_Layer}
 DisKCD constructs a heterogeneous relation-aware graph via relations between students, exercises, TKCs, and UKCs, linking four types of entities through a heterogeneous attention mechanism. So that each entity node can iteratively obtain updated information from neighbor nodes based on the attention weights.

\subsubsection{Heterogeneous Relation Attention.}
The heterogeneous relation attention occurs in a heterogeneous graph with five subgraphs, $G=(G_{se}$, $G_{ek}$, $G_{kk}$, $G_{ku}$, $G_{uu})$. For example, in $G_{uu}$ for UKCs there are prerequisite and similarity relations among them. Under both relations, each node contains different neighbor nodes. Specifically, after the Embedding Layer, we obtain the initial embeddings for four types of entities: $\boldsymbol{s}$, $\boldsymbol{e}$, $\boldsymbol{k}$ and $\boldsymbol{u}$. Within subgraphs $g\in G$, we exemplify with entity node $t\in(S,E,TKCs,UKCs)$. For entity node $t$, we calculate its attention weights with neighbor nodes under various relation contexts. Through weighted summation, we then derive the vector embedding of entity node $t$ in a subgraph $g$ after separately fusing different relations (In the subgraphs $G_{kk}$,$G_{ku}$, and $G_{uu}$, each one encompasses multiple types of relations). The implementation of heterogeneous relation attention is executed as follows:
\begin{equation}\label{eq:n(l+1)}
    {\widetilde{\boldsymbol{n}}}^{(l+1)}=\sum_{r\in R_g}\sum_{n\in N_t^r}{\alpha_n^{(l+1)}\boldsymbol{W}^{(l)}\boldsymbol{n}^{(l)}},
\end{equation}
\begin{equation}
    {\hat{\alpha}}_n^{\left(l+1\right)}=FNN([\boldsymbol{W}^{(l)}\boldsymbol{t}^{(l)},\boldsymbol{W}^{(l)}\boldsymbol{n}^{(l)}]),
\end{equation}
\begin{equation}\label{eq:alpha(l+1)}
    \alpha_n^{(l+1)}=\frac{exp({\hat{\alpha}}_n^{\left(l+1\right)})}{\sum_{j\in N_t^r}{exp({\hat{\alpha}}_j^{\left(l+1\right)})}},
\end{equation}
where $R_{g}$ denotes the set of relationship types between nodes in the heterogeneous subgraph $g$.
$N_{t}^{r}$denotes the set of neighboring nodes of node $t$ under the relationship $r$.
$\alpha_n^{(l+1)} $ refers to the attention weights assigned to node $t$'s neighbors, reflecting the significance of each relation type $r$ in the heterogeneous subgraph. $\boldsymbol{W}^{(l)}$ is the trainable matrix. $\boldsymbol{n}^{(l)}$ represents the embedding of a neighboring node of $t$ after the $l$-th iteration.
In the Heterogeneous Relation Attention module, we obtain the embedding representations of entities under various relations within a specific subgraph. 

In the next section, we will aggregate the embeddings of an entity from various subgraphs.

\subsubsection{Heterogeneous Relation Aggregation.}
In this section, we explore how to aggregate an entity's embeddings from various subgraphs, as an entity node may appear in multiple subgraphs. For example, node $u$ is present in both $G_{ku}$ and $G_{uu}$, serving as distinct roles. 
Since entities obtain different embeddings in each subgraph, for an entity node $t\in(S,E,TKCs,UKCs)$, the aggregated embeddings from the neighboring nodes in different subgraphs are concatenated with the $l$-th iteration embedding of node $t$ to generate the $l+1$-th iteration embedding of the current entity node. The aggregation and iterative process for each type of entity node is described as follows.
The aggregation iteration process for each class of entity nodes is as follows.

\textbf{Student Aggregation.}
The student nodes exist only in the student-exercise interaction subgraph $G_{se}$, so there is no need to fuse information from multiple subgraphs. The information aggregation process for the student nodes in this subgraph is shown as Eq.\ref{eq:n(l+1)} to Eq.\ref{eq:alpha(l+1)}. However, to enhance stability during training, we also incorporate the residual from the previous iteration into the embedding of student $s$ after the $(l+1)$-th iteration.
\begin{equation}
s^{(l+1)}=s^{(l)}+\widetilde{e}_s^{(l+1)}.
\end{equation}

Here, $\widetilde{e}_s^{(l+1)}$ represents the embedding vector of the student entity obtained through  Eq.\ref{eq:n(l+1)} to Eq.\ref{eq:alpha(l+1)}, $s^{(l)}$ represents the residual from the previous iteration.

\textbf{Exercise Aggregation.}
Exercises are present in both the student-exercise interaction subgraph and the exercise-TKC correlation subgraph. In the student-exercise interaction subgraph $G_{se}$, the aggregated representation of the neighboring nodes of exercise $e$ is denoted as $\widetilde{s_e}^{(l+1)}$, in the exercise-TKC related subgraph $G_{ek}$ the aggregated representation of the neighboring nodes of exercise $e$ is denoted as $\widetilde{k_e}^{(l+1)}$. The aggregation methods for $\widetilde{s_e}^{(l+1)}$ and $\widetilde{k_e}^{(l+1)}$ follow Eq.\ref{eq:n(l+1)}.

Finally, the embedding of exercise $e$ after the ${(l+1)}^{th}$ iteration is
\begin{equation}
    e^{(l+1)}=e^{(l)}+\mu_s^{(l+1)} \widetilde{s}_e^{(l+1)}+\mu_k^{(l+1)} \widetilde{k}_e^{(l+1)},
\end{equation}
\begin{equation}\label{eq:miuweight}
    \mu_s^{(l+1)}=FNN([e^{(l)},\widetilde{s}_e^{(l+1)}]),
\end{equation}
\begin{equation}
    \mu_k^{(l+1)}=FNN([e^{(l)},\widetilde{k}_e^{(l+1)}]).
\end{equation}
This process involves two subgraphs, with $\mu_s^{(l+1)}$ and $\mu_k^{(l+1)}$ representing the heterogeneous relation aggregation weights for the respective subgraphs.

\textbf{TKC Aggregation.}
A TKC can appear in three kinds of subgraphs: the exercise-TKC correlation subgraph, TKC-TKC dependency subgraph, and TKC-UKC dependency subgraph. In the exercise-TKC related subgraph $G_{ek}$, the aggregated representation of the neighboring nodes of TKC is denoted as $\widetilde{e}_k^{(l+1)}$. In the TKC-TKC dependency subgraph $G_{kk}$, the aggregated representation of the neighboring nodes of TKC is denoted as $\widetilde{k}_k^{(l+1)}$. In the TKC-UKC dependency subgraph $G_{ku}$, the aggregated representation of the neighboring nodes of TKC is denoted as $\widetilde{u}_k^{(l+1)}$. The aggregation methods for $\widetilde{e}_k^{(l+1)}$, $\widetilde{k}_k^{(l+1)}$ and $\widetilde{u}_k^{(l+1)}$ follow Eq.\ref{eq:n(l+1)}.

Finally, the embedding of the currently TKC $k$ after the ${(l+1)}^{th}$ iteration is 
\begin{equation}
    k^{(l+1)}=k^{(l)}+\mu_e^{(l+1)} \widetilde{e}_k^{(l+1)}+\mu_k^{(l+1)} \widetilde{k}_k^{(l+1)}+\mu_u^{(l+1)} \widetilde{u}_k^{(l+1)}.
\end{equation}
This process involves three kinds of maps, and $\mu_e^{(l+1)}$, $\mu_k^{(l+1)}$, and $\mu_u^{(l+1)}$ represent the heterogeneous relation aggregation weights of these subgraphs, respectively. The calculation method for the weights is similar to Eq.\ref{eq:miuweight}.

\textbf{UKC Aggregation.}
A UKC appears in two subgraphs: TKC-UKC dependency subgraph and UKC-UKC dependency subgraph. In the TKC-UKC dependency subgraph $G_{ku}$, the neighboring nodes of UKC are all TKCs related to it, and the aggregation of these neighboring nodes can be denoted as $\widetilde{k}_u^{(l+1)}$. In the UKC-UKC dependency subgraph, the aggregation of neighbor nodes of the current UKC node is $\widetilde{u}_u^{(l+1)}$. The aggregation methods for  $\widetilde{k}_u^{(l+1)}$ and $\widetilde{u}_u^{(l+1)}$ follow Eq.\ref{eq:n(l+1)}.

Finally, the embedding of the current UKC $u$ after the ${(l+1)}^{th}$ iteration is
\begin{equation}
    u^{(l+1)}=u^{(l)}+\mu_k^{(l+1)} \widetilde{k}_u^{(l+1)}+\mu_u^{(l+1)} \widetilde{u}_u^{(l+1)}.
\end{equation}
Here, $\mu_k^{(l+1)}$ and $\mu_u^{(l+1)}$ represent the heterogeneous relation aggregation weights of the two subgraphs, respectively. The calculation method for these weights is similar to Eq.\ref{eq:miuweight}
After aggregating from different subgraphs, we obtain the final feature representations of four types of entities: $\boldsymbol{s}^{(l+1)}$,$\boldsymbol{e}^{(l+1)}$,$\boldsymbol{k}^{(l+1)}$ and $\boldsymbol{u}^{(l+1)}$. All the feature vectors of these entities fully aggregate information from neighbouring nodes from different subgraphs, providing a solid foundation for the next step of student performance prediction. Next, we input the entity feature vectors generated by DisKCD into the generic CDM to examine students' proficiency in UKCs.

\subsection{Diagnosis Layer}
\label{ssec_DiagnosisLayer}

In this paper, we combine the proposed DisKCD framework with the interaction functions of several commonly used CDMs (including IRT, MIRT, DINA, NeuralCD, and RCD) to predict students' performance on exercises related to UKCs, thereby assessing their proficiency in UKCs. The codes for IRT, MIRT, and DINA are sourced from a GitHub project, while the codes for NeuralCD and RCD are obtained from their original literature, as referenced in \citep{DBLP:conf/aaai/WangLCHCYHW20} and \citep{DBLP:conf/sigir/Gao0HYBWM0021}.

Given that distinct CDM employ varying formats of student, exercise, and knowledge concept features as input, transform functions to convert the embeddings into the formats suitable for each CDM are necessary. Hence, we redefine Eq.\ref{eq:fcdm} as follows: 
\begin{equation}
    {\hat{y}}_{uv}=F_{CDM}(\phi_s(\boldsymbol{u}),\phi_e(\boldsymbol{v}),\phi_c(\boldsymbol{K})).
\end{equation}

Here, $\phi_s(\cdot)$, $\phi_e(\cdot)$, and $\phi_c(\cdot)$ represent transform functions. We utilize cross-entropy loss function to calculate the difference between the predicted probability ${\hat{y}}_{uv}$ and the actual response $y_{uv}$. 
\begin{equation}
    L=-\sum_{(s_u,e_v,y_{uv})\in R_{se}}{(y_{uv}{log{\hat{y}}}_{uv}+(1-y_{uv})log{(}1-{\hat{y}}_{uv}))}.
\end{equation}

\section{Experiments}
\label{sec_Experiments}

We conduct comprehensive experiments to answer the following research questions:

\textbf{RQ1}: Does DisKCD enhance the performance of CDMs in assessing students' mastery of UKCs?

\textbf{RQ2}: Is DisKCD beneficial for CDMs in evaluating students' mastery of TKCs?

\textbf{RQ3}: How well does DisKCD model entities like students, exercises, and knowledge concepts?

\textbf{RQ4}: Are the cognitive diagnosis results produced by DisKCD interpretable?

\subsection{Datasets}
We evaluated our approach on five real-world datasets: two from online learning platforms, Junyi\citep{DBLP:conf/edm/ChangHC15} and ASSISTments09\citep{wang2024survey}, which reflect students' proficiency in UKCs within online education environments, and three from standardized paper-and-pencil tests, namely JAD, SDP, and Math2\citep{hendrycks2021measuring}. In the paper-and-pencil test datasets, each student answers the same set of exercises, enabling us to diagnose students' proficiency in UKCs within a traditional offline classroom setting. By combining both online and offline datasets, we aim to comprehensively assess the diagnostic power of our model across different learning environments, thereby mitigating the limitations of relying solely on online course data. Details are shown in Table~\ref{tab:table_1}.

\begin{table}[width=1\linewidth,cols=5,pos=h]
\caption{The statistics of five datasets.}
\renewcommand{\arraystretch}{1.2}
\setlength{\tabcolsep}{2pt}
\label{tab:table_1}
\begin{tabular*}{\tblwidth}{@{} LLLLLL@{} }
\toprule
Dataset & $\#$Student & $\#$Exercise & $\#$Record & $\#$Record per student \\
\midrule
JAD & 275 & 18 & 4,950 & 18 \\
SDP & 128 & 15 & 1,920 & 15 \\
Junyi & 10,000 & 835 & 353,835 & 35.38 \\
ASSIST & 2,493 & 17,746 & 257,415 & 107.27 \\
Math2 & 3,911 & 20 & 78,220 & 20 \\
\bottomrule
\end{tabular*}
\end{table}

\textbf{JAD} and \textbf{SDP} consist of student response data extracted from the final exams of two courses, "Java Application Development" and "Software Design Patterns", offered during the 2021-2022 academic year in the software engineering major at a Chinese university. Note that the exercises in these exams are open questions rather than fill-in-the-blank questions or multiple-choice questions.

\textbf{Junyi} is sourced from the Chinese e-learning platform Junyi Academy, which is widely applied in CDM. To more effectively differentiate between TKCs and UKCs, we extracted response logs of 10,000 students in the Junyi dataset for our experiments.

\textbf{ASSISTments2009} collects from the ASSISTments platform during the 2009-2010 school year contains student practice data on questions associated with specific knowledge concepts. We use the refined version of this dataset, which resolves issues in earlier versions, to ensure reliability in our experiments.

\textbf{Math2} consists of data collected from two final high school math exams, including both objective and subjective problems, representing a standard test environment.

\begin{table*}[width=1\textwidth,pos=h] 
  
\caption{Performance comparison on Untested Knowledge Concept. The best performance prediction is highlighted in bold.}
\renewcommand{\arraystretch}{1.3}
\setlength{\tabcolsep}{2pt} 
\resizebox{\textwidth}{!}{ 
\begin{tabular}{llccccccccccclccclccc}
\hline
\multicolumn{2}{c}{Dataset}                          & \multicolumn{3}{c}{JAD}                          &  & \multicolumn{3}{c}{SDP}                          &  & \multicolumn{3}{c}{Junyi}                        &  & \multicolumn{3}{c}{ASSISTments09}                &  & \multicolumn{3}{c}{Math2}                        \\ \cline{3-5} \cline{7-9} \cline{11-13} \cline{15-17} \cline{19-21} 
\multicolumn{2}{c}{Methods}                                 & ACC            & RMSE           & AUC            &  & ACC            & RMSE           & AUC            &  & ACC            & RMSE           & AUC            &  & ACC            & RMSE           & AUC            &  & ACC            & RMSE           & AUC            \\ \hline
{\multirow{2}{*}{DINA}} & baseline & 0.594          & 0.543          & 0.594          &  & 0.578          & 0.512          & 0.568          &  & 0.608          & 0.474          & 0.652          &  & 0.625          & 0.502          & 0.502          &  & 0.554          & 0.568          & 0.526          \\
\multicolumn{1}{c}{}                      & DisKCD   & \textbf{0.641} & \textbf{0.493} & \textbf{0.648} &  & \textbf{0.643} & \textbf{0.497} & \textbf{0.646} &  & \textbf{0.750} & \textbf{0.431} & \textbf{0.805} &  & \textbf{0.650} & \textbf{0.499} & \textbf{0.525} &  & \textbf{0.572} & \textbf{0.536} & \textbf{0.560} \\ \hline
\multirow{2}{*}{IRT}                      & baseline & 0.592          & 0.538          & 0.598          &  & 0.526          & 0.572          & 0.512          &  & 0.646          & 0.508          & 0.724          &  & 0.615          & \textbf{0.471} & 0.499          &  & 0.496          & 0.519          & 0.519          \\
                                          & DisKCD   & \textbf{0.593} & \textbf{0.494} & \textbf{0.709} &  & \textbf{0.568} & \textbf{0.539} & \textbf{0.632} &  & \textbf{0.755} & \textbf{0.409} & \textbf{0.815} &  & \textbf{0.647} & 0.481          & \textbf{0.501} &  & \textbf{0.529} & \textbf{0.499} & \textbf{0.522} \\ \hline
\multirow{2}{*}{MIRT}                     & baseline & 0.534          & 0.542          & 0.631          &  & 0.549          & 0.554          & 0.645          &  & 0.654          & 0.535          & 0.726          &  & 0.631          & 0.471          & 0.519          &  & 0.534          & 0.514          & 0.500          \\
                                          & DisKCD   & \textbf{0.599} & \textbf{0.515} & \textbf{0.632} &  & \textbf{0.589} & \textbf{0.506} & \textbf{0.666} &  & \textbf{0.763} & \textbf{0.407} & \textbf{0.813} &  & \textbf{0.655} & \textbf{0.468} & \textbf{0.562} &  & \textbf{0.552} & \textbf{0.504} & \textbf{0.529} \\ \hline
\multirow{2}{*}{NeuralCD}                      & baseline & 0.590          & 0.576          & 0.604          &  & 0.526          & 0.581          & 0.577          &  & 0.705          & 0.429          & \textbf{0.789} &  & 0.645          & 0.491          & 0.501          &  & 0.533          & 0.520          & 0.413          \\
                                          & DisKCD   & \textbf{0.653} & \textbf{0.470} & \textbf{0.694} &  & \textbf{0.623} & \textbf{0.560} & \textbf{0.604} &  & \textbf{0.748} & \textbf{0.411} & 0.701          &  & \textbf{0.656} & \textbf{0.480} & \textbf{0.627} &  & \textbf{0.554} & \textbf{0.518} & \textbf{0.525} \\ \hline
\multirow{2}{*}{RCD}                      & baseline & 0.591          & 0.489          & 0.629          &  & 0.607          & 0.496          & 0.658          &  & 0.628          & 0.477          & 0.646          &  & 0.655          & 0.466          & 0.652          &  & 0.547          & 0.511          & 0.593          \\
                                          & DisKCD   & \textbf{0.667} & \textbf{0.460} & \textbf{0.695} &  & \textbf{0.677} & \textbf{0.450} & \textbf{0.755} &  & \textbf{0.775} & \textbf{0.394} & \textbf{0.772} &  & \textbf{0.675} & \textbf{0.465} & \textbf{0.741} &  & \textbf{0.576} & \textbf{0.500} & \textbf{0.594} \\ \hline
\end{tabular}
}
\label{tab:table_2}
\end{table*}

\subsection{Experiment Settings}

\textbf{Baselines.} To evaluate the validity and generalizability, we employed DisKCD with five widely adopted cognitive diagnostic models (CDMs):
\begin{itemize}
\item DINA \citep{de2009dina} is a Q-matrix-based model that evaluates knowledge mastery by assuming correct answers only if all required skills are mastered.
\item IRT \citep{lord1952theory} is a probabilistic model that assesses cognitive status by analyzing item parameters such as difficulty and discrimination.
\item MIRT \citep{Reckase2009} is an extension of IRT, allowing multiple dimensions to represent student abilities and item characteristics across various skills.
\item NeuralCD \citep{DBLP:conf/aaai/WangLCHCYHW20} is a neural network-based model that captures higher-order interactions between students and items using multilayer perceptrons.
\item RCD \citep{DBLP:conf/sigir/Gao0HYBWM0021} constructs relational maps among students, items, and concepts to enhance cognitive diagnosis with multi-layered connections.
\end{itemize}






\textbf{Metrics.} Following previous studies, we used various metrics from regression and classification perspectives. From the regression perspective, we selected Root Mean Square Error (RMSE) \citep{DBLP:journals/corr/abs-1712-00328} to measure the difference between predicted and actual scores. From the classification perspective, we used prediction Accuracy (ACC) \citep{DBLP:conf/aaai/WangLCHCYHW20} and Area Under an ROC Curve (AUC) \citep{DBLP:journals/pr/Bradley97} for model evaluation.

\textbf{Implementation Details.} We set the dimensions of the student and exercise vectors to equal the number of unique knowledge concepts in each course, aligning with the structure of the underlying Q-matrix. All network parameters were initialized using Xavier initialization to maintain variance across layers and promote stable training. For training, the batch size was set according to dataset size and complexity: 8 for JAD, SDP, and Math2 datasets to prevent overfitting on smaller datasets, and 128 for the larger Junyi and ASSISTments2009 datasets to enable faster convergence. We optimized the model using the Adam optimizer with a learning rate of 0.001 to balance the convergence speed and stability. To support interaction functions in the IRT and MIRT models, we included additional parameters: one to control the range of item difficulty and discrimination scores, which was set to a maximum of 1 to ensure these values stay within a consistent, interpretable range. Another parameter, set to 5, determines the number of dimensions in the latent trait space, allowing the interaction functions to capture more complex relationships between student abilities and question characteristics. The code was implemented in PyTorch, and all experiments were run on a Linux server equipped with an RTX 4090 GPU, enabling efficient handling of large-scale computations.

\begin{table*}[width=1\textwidth,pos=h]
\caption{Performance comparison on Tested Knowledge Concept.}
\renewcommand{\arraystretch}{1.3}
\setlength{\tabcolsep}{2pt} 
\resizebox{\textwidth}{!}{ 
\begin{tabular}{llccccccccccclccclccc}
\hline
\multicolumn{2}{c}{Dataset}      & \multicolumn{3}{c}{JAD}                          &  & \multicolumn{3}{c}{SDP}                          &  & \multicolumn{3}{c}{Junyi}                        &  & \multicolumn{3}{c}{ASSISTments09}                & \multicolumn{1}{c}{} & \multicolumn{3}{c}{Math2}                        \\ \cline{3-5} \cline{7-9} \cline{11-13} \cline{15-17} \cline{19-21} 
\multicolumn{2}{c}{Methods}             & ACC            & RMSE           & AUC            &  & ACC            & RMSE           & AUC            &  & ACC            & RMSE           & AUC            &  & ACC            & RMSE           & AUC            &                      & ACC            & RMSE           & AUC            \\ \hline
\multirow{2}{*}{DINA} & baseline & 0.709          & 0.449          & 0.701          &  & 0.617          & 0.496          & 0.686          &  & 0.641          & 0.474          & 0.754          &  & 0.684          & 0.473          & \textbf{0.724} &                      & 0.586          & 0.504          & 0.700          \\
                      & DisKCD   & \textbf{0.752} & \textbf{0.437} & \textbf{0.721} &  & \textbf{0.659} & \textbf{0.481} & \textbf{0.712} &  & \textbf{0.672} & \textbf{0.458} & \textbf{0.802} &  & \textbf{0.697} & \textbf{0.467} & 0.705          &                      & \textbf{0.597} & \textbf{0.490} & \textbf{0.736} \\ \hline
\multirow{2}{*}{IRT}  & baseline & 0.750          & 0.432          & 0.643          &  & 0.682          & 0.471          & 0.710          &  & 0.646          & 0.508          & 0.795          &  & 0.685          & 0.438          & 0.755          &                      & 0.716          & \textbf{0.435} & 0.782          \\
                      & DisKCD   & \textbf{0.764} & \textbf{0.408} & \textbf{0.732} &  & \textbf{0.711} & \textbf{0.464} & \textbf{0.729} &  & \textbf{0.755} & \textbf{0.409} & \textbf{0.812} &  & \textbf{0.714} & \textbf{0.436} & \textbf{0.755} &                      & \textbf{0.720} & 0.437          & \textbf{0.786} \\ \hline
\multirow{2}{*}{MIRT} & baseline & 0.707          & 0.447          & 0.608          &  & 0.690          & 0.469          & 0.712          &  & 0.763          & 0.407          & 0.813          &  & 0.728          & 0.427          & 0.758          &                      & 0.726          & 0.428          & 0.784          \\
                      & DisKCD   & \textbf{0.770} & \textbf{0.414} & \textbf{0.756} &  & \textbf{0.701} & \textbf{0.462} & \textbf{0.741} &  & \textbf{0.771} & \textbf{0.398} & \textbf{0.827} &  & \textbf{0.730} & \textbf{0.424} & \textbf{0.766} &                      & \textbf{0.732} & \textbf{0.426} & \textbf{0.785} \\ \hline
\multirow{2}{*}{NeuralCD}  & baseline & 0.740          & 0.415          & 0.710          &  & 0.628          & 0.479          & 0.669          &  & 0.739          & 0.417          & 0.796          &  & 0.725          & 0.431          & 0.749          &                      & 0.720          & 0.448          & 0.733          \\
                      & DisKCD   & \textbf{0.751} & \textbf{0.404} & \textbf{0.753} &  & \textbf{0.663} & \textbf{0.478} & \textbf{0.698} &  & \textbf{0.750} & \textbf{0.411} & \textbf{0.803} &  & \textbf{0.729} & 0.431 & \textbf{0.758} &                      & \textbf{0.721} & \textbf{0.439} & \textbf{0.752} \\ \hline
\multirow{2}{*}{RCD}  & baseline & 0.759          & 0.405          & 0.731          &  & 0.661          & 0.459          & 0.722          &  & 0.769          & 0.398          & 0.826          &  & 0.729          & \textbf{0.423} & 0.764          &                      & 0.729          & 0.427          & 0.785          \\
                      & DisKCD   & \textbf{0.776} & \textbf{0.401} & \textbf{0.739} &  & \textbf{0.712} & \textbf{0.457} & \textbf{0.740} &  & \textbf{0.773} & \textbf{0.396} & \textbf{0.830} &  & \textbf{0.773} & 0.425          & \textbf{0.768} &                      & \textbf{0.732} & \textbf{0.425} & \textbf{0.786} \\ \hline
\end{tabular}
}
\label{tab:table_3}
\end{table*}

\begin{table*}[width=1\textwidth,pos=h]
\caption{The performance of entity modeling.}
\renewcommand{\arraystretch}{1.2}
\setlength{\tabcolsep}{2pt}
\label{tab:table_4}
\centering
\resizebox{\textwidth}{!}{

\begin{tabular}{cccccccccccc}
\hline
Methods                        & \multirow{2}{*}{Metric} & \multicolumn{2}{c}{DINA} & \multicolumn{2}{c}{IRT}  & \multicolumn{2}{c}{MIRT} & \multicolumn{2}{c}{NeuralCD} & \multicolumn{2}{c}{RCD}  \\ \cline{3-12} 
Dataset                        &                         & w/o HRL & OURS           & w/o HRL & OURS           & w/o HRL & OURS           & w/o HRL   & OURS             & w/o HRL & OURS           \\ \hline
\multirow{3}{*}{JAD}           & ACC                     & 0.591   & \textbf{0.641} & 0.592   & \textbf{0.593} & 0.543   & \textbf{0.599} & 0.590     & \textbf{0.653}   & 0.627   & \textbf{0.667} \\
                               & RMSE                    & 0.544   & \textbf{0.493} & 0.538   & \textbf{0.494} & 0.554   & \textbf{0.515} & 0.509     & \textbf{0.470}   & 0.477   & \textbf{0.460} \\
                               & AUC                     & 0.589   & \textbf{0.648} & 0.534   & \textbf{0.709} & 0.547   & \textbf{0.632} & 0.637     & \textbf{0.694}   & 0.665   & \textbf{0.695} \\ \hline
\multirow{3}{*}{ASSISTments09} & ACC                     & 0.646   & \textbf{0.697} & 0.645   & \textbf{0.647} & 0.645   & \textbf{0.655} & 0.645     & \textbf{0.656}   & 0.636   & \textbf{0.675} \\
                               & RMSE                    & 0.502   & \textbf{0.467} & 0.493   & \textbf{0.481} & 0.469   & \textbf{0.468} & 0.486     & \textbf{0.480}   & 0.471   & \textbf{0.465} \\
                               & AUC                     & 0.515   & \textbf{0.705} & 0.488   & \textbf{0.501} & 0.526   & \textbf{0.562} & 0.581     & \textbf{0.627}   & 0.557   & \textbf{0.741} \\ \hline
\end{tabular}
}
\end{table*}

\subsection{Performance Comparison on UKCs (RQ1)}

To investigate RQ1, we trained our models using data related to TKCs and made predictions on data related to UKCs. We utilized interaction functions from five popular CDMs, comparing the original models with those that use feature vectors generated through DisKCD. The results, as presented in Table~\ref{tab:table_2}, led to the following observations:
(1) In nearly all CDMs, vectors updated through DisKCD perform better than those updated through the original models, with consistent improvements across ACC and AUC metrics. Specifically, DisKCD showed a 2\% to 6\% improvement over baseline CDM on the metric ACC, suggesting that DisKCD effectively captures critical knowledge relations, thus improving the accuracy of diagnosing students' proficiency on UKCs.
(2) Across the five datasets, DisKCD generally outperforms the baseline across various metrics, highlighting the advantage of feature vectors generated using heterogeneous relational graph networks. The Junyi dataset, in particular, exhibits substantial gains, with DisK-IRT improving ACC from 0.646 to 0.755 and AUC from 0.795 to 0.812. This emphasizes DisKCD's ability to address sparse data challenges, particularly for datasets where traditional CDMs struggle with limited UKCs information.
(3) The experimental results show that the DisK-RCD model generally outperforms DisK-DINA and DisK-IRT, highlighting the effectiveness of modeling complex student-exercise-concept relationships. The performance boost provided by DisKCD on these models further validates this point. For example, on the JAD dataset, the ACC of DisK-RCD reached 0.776, while DisK-DINA and DisK-IRT only achieved ACC values of 0.752 and 0.764, respectively. These findings emphasize the importance of complex student-exercise-concept interactions in diagnosing students' proficiency, suggesting that DisKCD can better leverage such intricate relationships to provide more accurate assessments.

These performance improvements can be attributed to DisKCD’s ability to leverage Heterogeneous Relation Attention and Heterogeneous Relation Aggregation, which enables the model to deeply explore the relationships between entities and effectively disentangle TKCs from UKCs, capturing finer details of student knowledge mastery. By integrating these elements, DisKCD provides a comprehensive understanding of students' knowledge gaps, enhancing its overall effectiveness in diagnosing UKCs. In conclusion, the experimental findings from Table~\ref{tab:table_2} demonstrate that DisKCD is a robust and effective approach for diagnosing UKCs. By utilizing direct connections along with heterogeneous relation attention and aggregation mechanisms, DisKCD significantly improves feature extraction, allowing it to accurately capture students’ understanding across both directly and indirectly assessed concepts. The observed improvements underscore DisKCD’s potential as a valuable tool for real-time, adaptive learning systems and highlight its adaptability across different datasets and model types.

\begin{figure*}
  \centerline{\includegraphics[width=1\textwidth]{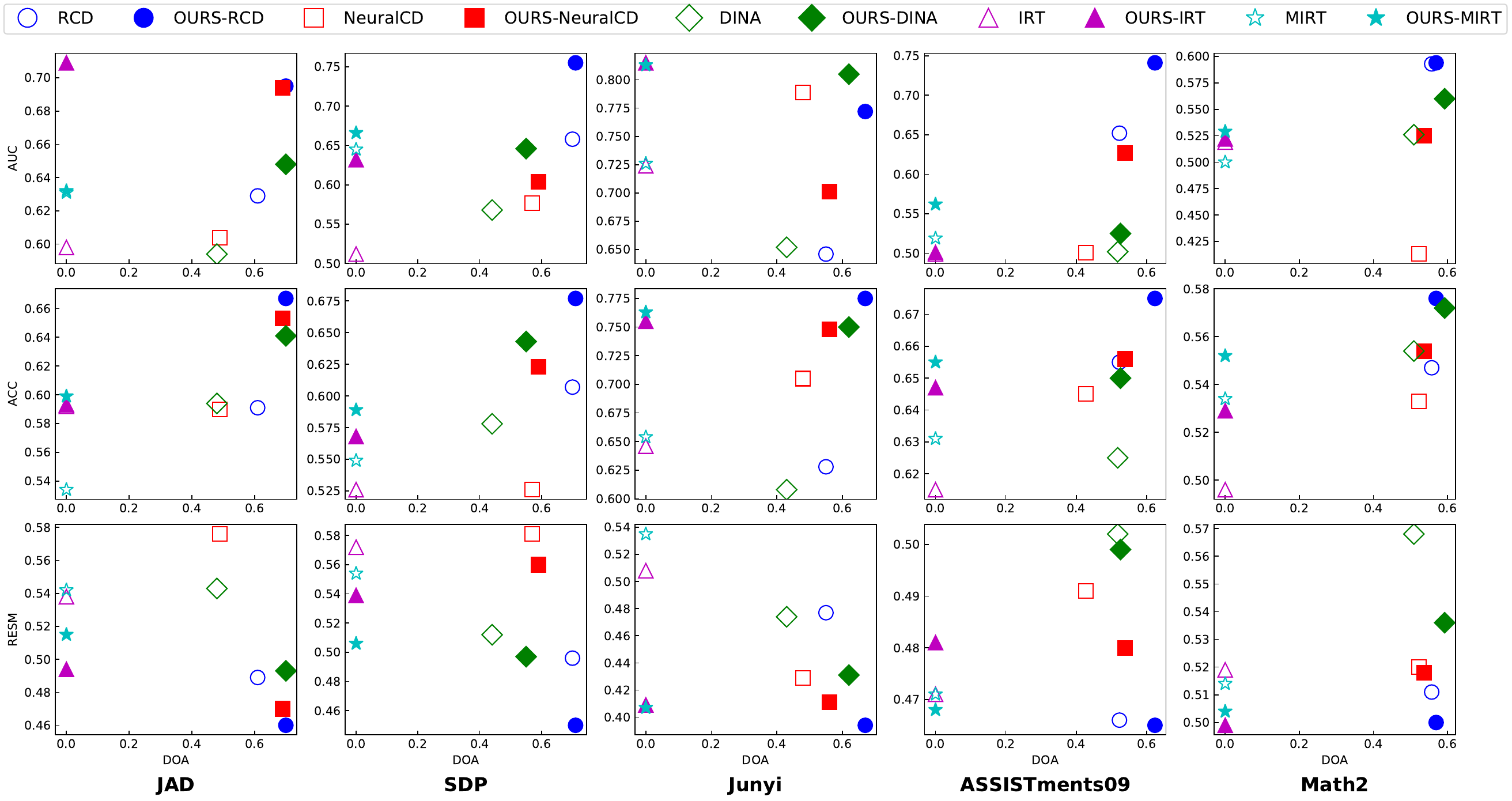}}
  \caption{Overview of AUC, ACC, RESM, and interpretability performance of DisKCD and the baseline across five datasets.}\label{FIG:3}
\end{figure*}

\subsection{Performance Comparison on TKCs (RQ2)}
In the experiments addressing RQ2, the exercises in both training and testing datasets were related to TKCs. As shown in Table~\ref{tab:table_3}, the diagnostic results based on feature vectors updated through DisKCD generally outperform those of the baseline model across all CDMs, indicating DisKCD's advantage in more accurately assessing students' mastery of TKCs. This improvement is attributed to DisKCD’s capacity to model the nuanced relationships within TKCs, providing a more precise evaluation of student knowledge. (1) Notably, the DisK-RCD model showed significant increases in ACC and AUC, underscoring DisKCD's effectiveness in capturing complex interactions. (2) Experiments on the Junyi and ASSISTments09 datasets generally yielded higher AUC values compared to our custom datasets, JAD, SDP, and Math2, indicating that models trained on richer, more diverse data exhibit stronger predictive capabilities. The greater richness of the Junyi and ASSISTments09 datasets likely contributes to their enhanced predictive power. (3) Additionally, these findings emphasize the importance of models that can adapt to different types of knowledge concepts. DisKCD effectively distinguishes and models concepts in tasks involving both frequently and less frequently assessed knowledge, offering a balanced approach for cognitive diagnosis. The consistent improvements across datasets further demonstrate DisKCD’s robustness and flexibility, suggesting its potential for application across varied educational contexts with diverse data availability.

\subsection{Ability of Modeling Entity Embedding (RQ3)}
To understand the influence of the Heterogeneous Relation-aware Layer within DisKCD on overall model performance, we conducted a series of comparative experiments on both the JAD and ASSISTments09 datasets. JAD represents a traditional paper-and-pencil test environment, while ASSISTments09 represents an online educational platform. We use the entity feature vectors obtained from one-hot vector-based learning as a baseline, representing DisKCD without the Heterogeneous Relation-aware Layer (w/o HRL), which relies on one-hot embeddings rather than the feature vectors generated by this layer.

As shown in Table~\ref{tab:table_4}, the feature vectors produced by DisKCD’s Heterogeneous Relation-aware Layer significantly improved diagnostic performance on both datasets, underscoring the importance of these representations in capturing nuanced entity relationships. This layer not only excels at modeling complex and heterogeneous relationships within educational data, but it also demonstrates robust adaptability across varied educational contexts. The consistent improvements observed across both traditional and online learning environments highlight the versatility of the Heterogeneous Relation-aware Layer, making it a valuable asset for enhancing diagnostic precision and flexibility in diverse educational settings.

\begin{figure*}
  \centerline{\includegraphics[width=1\textwidth]{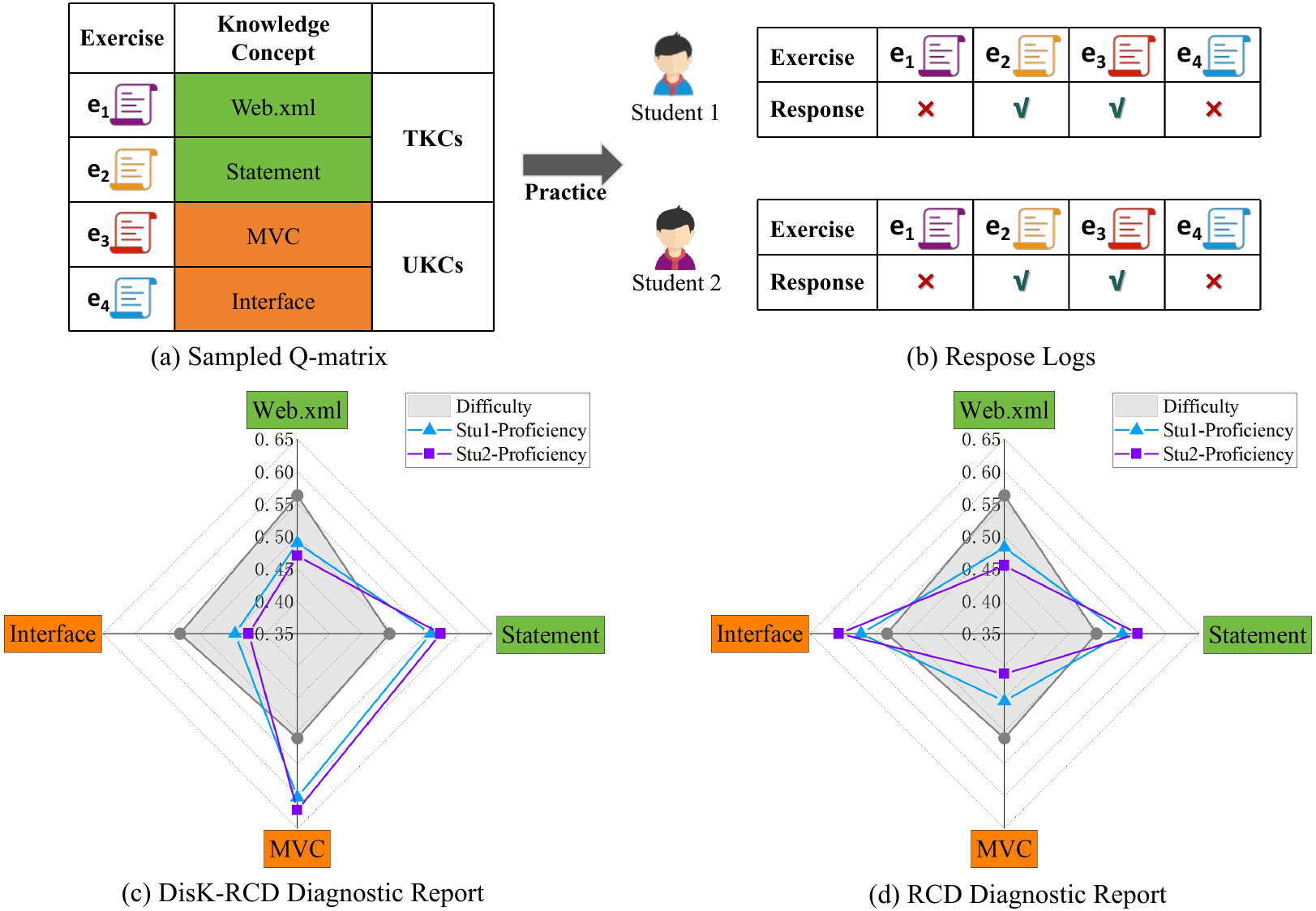}}
  \caption{Comparison of diagnostic results for two students under the DisK-RCD and RCD models.}\label{FIG:4}
\end{figure*}

\subsection{Interpretability and Case Study (RQ4)}

We evaluated the interpretability of DisKCD’s diagnostic results using the Degree of Agreement (DOA), as proposed by \citep{4072747}, This metric is rooted in the Monotonicity Assumption established by \citep{Reckase2009}, which suggests that students with higher proficiency in a concept should perform better on related exercises. The DOA formula is as follows. 

\begin{equation}
\scalebox{0.85}{$
DOA(i) = \frac{1}{Z} \displaystyle \sum_{a=1}^{N} \sum_{b=1}^{N} \delta(P_{a,i},P_{b,i}) \sum_{j=1}^{M} I_{j,i} \frac{J(j,a,b) \wedge \delta(r_{aj},r_{bj})}{J(j,a,b)}
$},
\end{equation}
where $\mathrm{Z=}\sum_{a=1}^{N}\sum_{b=1}^{N}\delta\left(P_{a,i},P_{b,i}\right)$, $P_{a,i}$ and $P_{b,i}$ denote student  $s_a$ and $s_b$ mastery of concept $k_i$, If $P_{a,i} > P_{b,i} , \delta\left(P_{a,i},P_{b,i}\right)=1$, 0 otherwise.  $I_{j,i}=1$ if $e_j$ contains knowledge concept $k_i$  and $0$ otherwise. $J\left(j,a,b\right)$ if both students have answered exercise $e_j$ , $0$ otherwise.

To explore the interpretability of DisKCD, Fig~\ref{FIG:3} and Fig~\ref{FIG:4} provide complementary analyses: Fig~\ref{FIG:3} uses data points to highlight DisKCD’s diagnostic strengths, while Fig~\ref{FIG:4} offers a case study of specific student responses.

\textbf{Analyzing Interpretability through Data Points.} Fig~\ref{FIG:3} shows the diagnostic results for exercises related to UKCs across five different datasets. Solid nodes denote DisKCD, while hollow nodes denote the baseline. In all subplots, the horizontal coordinates represent DOA, and the vertical coordinates represent various metrics (ACC, AUC, RMSE). In the subplots for accuracy (ACC) and area under the curve (AUC), nodes positioned further to the top right indicate better performance, while in the subplots for root mean square error (RMSE), nodes positioned further to the bottom right reflect better performance. As shown in Fig~\ref{FIG:3}, the higher DOA values of the solid nodes indicate that DisKCD’s diagnosis is more closely aligned with actual student performance. Additionally, solid nodes generally outperform hollow nodes on ACC, AUC, and RMSE metrics, further validating DisKCD’s diagnostic capabilities. It is worth noting that the latent vectors in Item Response Theory (IRT) and Multidimensional Item Response Theory (MIRT) have no direct correspondence with specific concepts, resulting in DOA values of zero.

\textbf{Case Study Analysis for Interpretability.} Fig~\ref{FIG:4}, based on an experiment using the JAD dataset, Panel (a) of Fig~\ref{FIG:4} illustrates the connection between exercises and both Tested Knowledge Components (TKCs), such as Web.xml and Statements, and Untested Knowledge Concepts (UKCs), including MVC and Interfaces. Panel (b) shows students’ responses, while Panels (c) and (d) compare the diagnostic results of two students using DisK-RCD (left) and the original RCD (right). Notably, students answered correctly only when their proficiency exceeded the difficulty of the exercise. The analysis reveals that for the tested knowledge points (Web.xml and Statement), both DisK-RCD and RCD accurately capture students’ proficiency levels and correctly predict response outcomes. However, for untested knowledge points (MVC and Interface), DisK-RCD successfully identifies students’ proficiency, while the original RCD model produces inaccurate proficiency estimates. This demonstrates the effectiveness of the embeddings generated by the DisKCD model in diagnosing untested knowledge components.

\section{Conclusions}
\label{sec_Conclusions}
In this paper, we propose a general knowledge diagnosis framework called DisKCD. Specifically, we define four types of entities: students, exercises, TKCs, and UKCs, along with their fine-grained relations. This allows us to disentangle TKCs and UKCs. Then, using a heterogeneous relation-aware network, we generate relation-aware entity embeddings, which can be used in various CDMs to predict students' performance. Experimental results on real-world datasets show that DisKCD not only effectively predicts students' proficiency on UKCs but also helps assess their mastery of TKCs, with good interpretability.

Building on these findings, future research will aim to explore more diverse educational dependencies to further enhance CDMs, improve inference of students' cognitive states, and customize personalized learning paths. We hope our work will contribute to the field of intelligent education by providing insights and laying the groundwork for more advanced, adaptive learning models.

\bibliographystyle{cas-model2-names}

\bibliography{cas-refs}

\end{sloppypar}
\end{document}